# The Placement of the Head that Maximizes Predictability. An Information Theoretic Approach

*Ramon Ferrer-i-Cancho*[1]

**Abstract:** The minimization of the length of syntactic dependencies is a well-established principle of word order and the basis of a mathematical theory of word order. Here we complete that theory from the perspective of information theory, adding a competing word order principle: the maximization of predictability of a target element. These two principles are in conflict: to maximize the predictability of the head, the head should appear last, which maximizes the costs with respect to dependency length minimization. The implications of such a broad theoretical framework to understand the optimality, diversity and evolution of the six possible orderings of subject, object and verb, are reviewed.

## 1. Introduction

When producing an utterance speakers have to arrange elements linearly, forming a sequence. The same problem applies to users of a sign language or unconventional gesture systems (Goldin-Meadow 1999). Suppose that we have to order linearly a head and its dependents (complements or modifiers). In a verbal sequence made of subject, verb and object, we assume that the verb is the head. In a gestural sequence made of actor, action and patient, we assume that the action is the head. In general, what is the best placement of the head?

For the particular case of the ordering of the verb (i.e. the head) and the subject and the object (i.e. the complements), various sources of evidence suggest a preference for placing the verb last. First, the non-verbal experiments in (Goldin-Meadow et al 2008, Langus & Nespor 2010) where a robust strong preference for an order consistent with subject-object-verb (head last) was found even in speakers whose language did not have subject-object-verb as the dominant word order. Second, *in silico* experiments with neural networks have shown that subject-object-verb (head last) is the word order that emerges when languages are selected to be more easily learned by networks predicting the next element in a sequence (Reali & Christiansen 2009). Thirdly, the most frequent dominant word order among world languages is subject-object-verb (head last) (Drier 2013, Hammarström 2016). Table 1 shows that the total frequency of dominant orders increases as the head (V) moves from the beginning of the sequence (VOS/VSO) to the center (SVO/OVS) and finally to the end

[1] Complexity and Quantitative Linguistics Lab. LARCA Research Group. Departament de Ciències de la Computació, Universitat Politècnica de Catalunya (UPC). Campus Nord, Edifici Omega, Jordi Girona Salgado 1-3. 08034 Barcelona, Catalonia (Spain). Phone: +34 934134028. E-mail: rferrericancho@cs.upc.edu.

(SOV/OSV). That fact suggests that postponing the verb (the head) is favored for some reason.

Table 1
The frequency of the placement of the ordering of the subject (S), verb (V) and object (O) in world languages showing a dominant word order. Frequency is measured in languages and in families.

| Order | Languages | | Families | |
|---|---|---|---|---|
| | Frequency | Percentage | Frequency | Percentage |
| SOV | 2275 | 43.3 | 239 | 65.3 |
| SVO | 2117 | 40.3 | 55 | 15.0 |
| VSO | 503 | 9.6 | 27 | 7.4 |
| VOS | 174 | 3.3 | 15 | 4.1 |
| OVS | 40 | 0.8 | 3 | 0.8 |
| OSV | 19 | 0.4 | 1 | 0.3 |
| No dominant order | 124 | 2.4 | 26 | 7.1 |
| **V | 2294 | 43.7 | 240 | 65.6 |
| *V* | 2157 | 41.1 | 58 | 15.8 |
| V** | 677 | 13.9 | 42 | 11.5 |
| All | 5252 | | 366 | |

**V is used for verb final orderings (SOV and OSV), *V* is used for central verb placements (SVO and OVS) and V** for verb initial orderings (VSO and VOS). Frequency is measured in languages and also in families. Absolute frequencies are borrowed from (Hammarström 2016). Percentages were rounded to the nearest decimal.

Here we will provide general information theoretic arguments that predict that the verb (or the head in general) should be postponed and eventually placed last to maximize its predictability. The outline of the argument is as follows. Consider two practically equivalent pressures: the minimization of the uncertainty about a target element, and the maximization of the predictability of a target element (they are equivalent for sequences of length three or longer as explained in detail in Section 2). A target element is a specific element of the sequence that has not been produced yet. For simplicity, suppose that the sequence consists of word forms and the target is a word form. This setup can be easily adapted to other contexts, e.g., in animal behavior research, the target could be a type (of behavior) and the sequence would be made of types (Section 2 presents a generalization of the setup). We may choose a target between a head and its dependents or between a verb and its arguments. These pressures predict that the target element should be placed last. This result is intuitive: adding more elements before the target element cannot hurt (a reduction in predictability would hurt), and in general will help to predict it or to reduce its uncertainty (Cover & Thomas 2006). Similarly, Fenk-Oczlon (1989) stated that, "*as a linguistic sequence progresses, the number of possible continuations becomes more and more restricted; that is, there is a reduction of uncertainty of the information*".

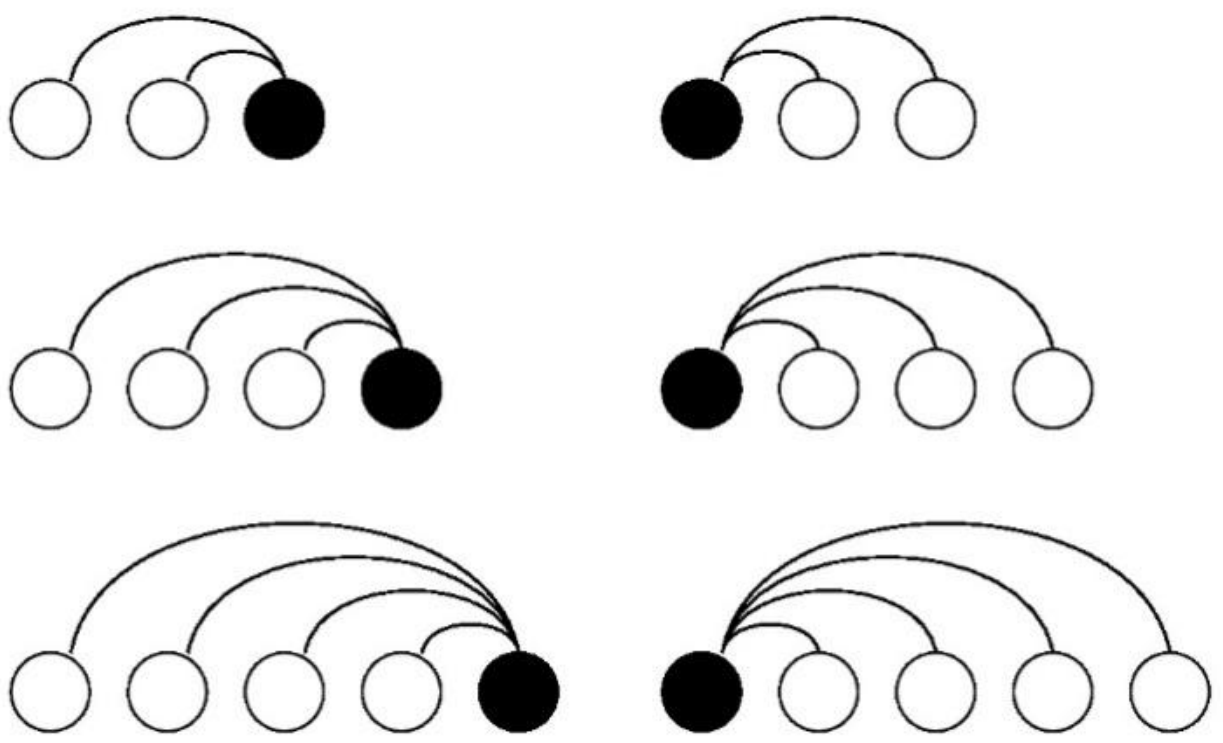

Figure 1. Optimal sequential placement of a head and its dependents (modifier/ complements) according to predictability maximization (or uncertainty minimizeation) for sequences of increasing length *m*.

The black circle indicates the head while the white circles indicate the dependents. Edges indicate syntactic dependencies between a head and its dependents. Top: $m = 3$. Center: $m = 4$. Bottom: $m = 5$. The left column indicates the optimal placements when the head is the target of predictability maximization. The right column corresponds to the optimal placement when the target are the dependents.

In case that the target element is the head, the result above implies that the head should be placed last (Fig. 1, left column). Assuming that the verb is the head the latter implies left branching. For the particular case of the subject-verb-object triple, the verb (or the action) then should be placed after the subject and the object (or the agent and the action). Interestingly, placing the verb at the center is not optimal but it is better than putting it first: postponing the verb is increasingly beneficial. In case that the target elements are the dependents, the head should be put first (Fig 1, right column). This implies right branching; the verb or the action should be the first element. The key is to understand why there should be a preference for the verb (or heads in general) to be the target.

These considerations notwithstanding, language is a multiconstraint engineering problem (Evans & Levinson 2009, Zipf 1949). Uncertainty minimization / predictability maximization are not the only relevant pressure in word order. An alternative well-established principle of word order is dependency length minimization (Liu 2017, Ferrer-i-Cancho 2015a). Suppose that we define the length of a dependency as the linear distance in words between the head and the dependent. If the head and the dependent are adjacent, the length is 1; if they are separated by one element, the length is 2; and so on…The principle of dependency length minimization consists of minimizing the sum of those dependencies. According to that principle, the optimal placement of a single head and its *n* dependents is at the center (Ferrer-i-Cancho 2015a). The length of the sequence is $m = n + 1$. If *m* is odd then there is only one possible central placement (Top and bottom of Fig. 2) while if it is even then there are two central placements (Center of Fig. 2). For this reason, the placement of the head is irrelevant when the se_quence only has two elements. The predictions of a central placement by the principle of dependency length minimization is exact if the dependents are atomic, i.e. made of just one word (Ferrer-i-Cancho 2015a), and is approximately valid when they are not (Ferrer-i-Cancho 2008, Ferrer-i-Cancho 2014). The argument can be refined (and generalized) supposing that the cognitive cost of a dependency increases as its length increases, and that the target of the minimization is the sum of the costs of all dependencies. Again, the optimal placement of the head is at the center (Ferrer-i-Cancho 2015a, Ferrer-i-Cancho 2014). The argument can also be refined measuring length in letters or phonemes instead of words (Ferrer-i-Cancho 2015b).

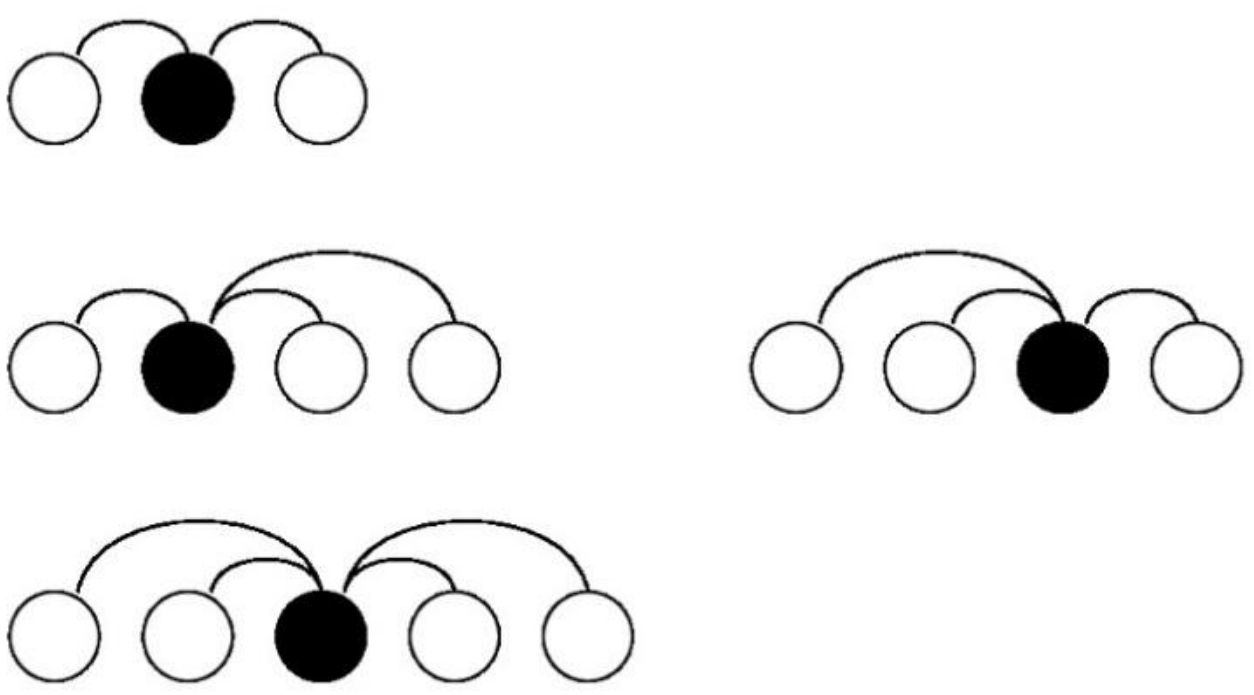

Figure 2. Optimal sequential placement of a head and its dependents according to dependency length minimization for sequences of increasing length *m*.

The black circle indicates the head while the white circles indicate the dependents. Edges indicate syntactic dependencies between a head and its modifier/complements. Top: $m = 3$ with only one optimal placement Center: $m = 4$ with two optimal placements. Bottom: $m = 5$ with only one optimal placement.

Interestingly, the principle of dependency length minimization is in conflict with the principle of predictability maximization / uncertainty minimization: while the former predicts that the head should be placed at the center of the sequence, the latter predicts that it should be placed at one of the ends. This article explores the implications of these conflicts and how they can be integrated into a general theory of word order.

The remainder of the article is organized as follows. Section 2 presents the mathematical arguments in detail. In our information theoretic approach, uncertainty is formalized as an entropy and predictability is formalized as a mutual information. We will show that uncertainty minimization has higher predictive power than mutual information maximization and we will also show that the former is equivalent to the latter for a sequence of at least three elements. This section is recommended to readers who lack the intuitions behind the results summarized above. Section 3 reviews the constant entropy rate and other information theoretic hypotheses since they are often regarded as reference theories. Section 4 presents a broad perspective on word order theory, incorporating the information theoretic approach elaborated in Section 2 and discussing implications for the ordering of subject, verb and object or its semantic correlates, i.e. actor, action and patient. Sections 2 and 3 can be skipped.

## 2. Information theory of word order

We aim to provide an information theoretic approach to word order that is consistent with other information theoretic approaches to language. Our guiding principle is that *"Scientific knowledge is systematic: a science is not an aggregation of disconnected information, but a system of ideas that are logically connected among themselves. Any system of ideas that is characterized by a certain set of fundamental (but refutable) peculiar hypotheses that try to fit a class of facts is a theory"* (Bunge, 2013, pp. 32-33).

For this reason, we will extend information theoretic principles that have been successful in explaining various linguistic phenomena: entropy minimization and mutual information maximization. A family of optimization models of natural communication is based on a combination of minimization of $H(S)$, the entropy of words of a vocabulary $S$, and the maximization of $I(S,R)$, the mutual information between the words ($S$) and the meanings (from a repertoire $R$). Here we extend and generalize this principles to be able to model word order phenomena. We refer the reader to Ferrer-i-Cancho (2017a) for a review of the cognitive and information theoretic justification of these principles. We also refer the reader to Chapter 2 of Cover & Thomas (2006) for further mathematical details about entropy, mutual information and conditional entropy.

We model a linguistic sequence (e.g. a sentence) as a sequence of elements $X_1$, $X_2$, $X_3$,...(e.g., the words of the sentence). First, let us consider $H(S)$. We proceed by replacing $S$ (a whole vocabulary) by a target of a sequence $Y$ and conditioning on elements of the sequence that have already appeared. This yields $H(Y|X_1,X_2, X_3,...)$. We postulate that this conditional entropy has to be minimized as $H(S)$. The next subsection presents the details of this minimization. We note that the minimization of entropy could be an axiom or a side-effect of compression. In the case of a vocabulary, the goal of compression is to minimize $L(S)$, the mean length of words. Interestingly, $L(S)$ is bounded below by $H(S)$ under the constraint of uniquely decipherability (Ferrer-i-Cancho, 2017a). Thus, minimizing $H(S)$ could be a consequence of pressure of the minimization of $L(S)$. The possibility that the minimization of $H(Y|X_1,X_2, X_3,...)$ is a side-effect of compression should be the subject of future research. The reason is that compression has the potential to offer a parsimonious explanation to various linguistic laws, including the popular Zipf's law for word frequencies (Ferrer-i-Cancho 2016b) and also Zipf's law of abbreviation (Ferrer-i-Cancho et al 2013b, Ferrer-i-Cancho et al 2015) and Menzerath's law (Gustison et al 2016).

Second, let us consider $I(S,R)$. As before, we proceed by replacing $S$ (a whole vocabulary) by a target of a sequence $Y$ and replacing $R$ by elements of the sequence that have already appeared. This yields $I(Y; X_1, X_2, X_3,...)$. We postulate that this mutual information has to be maximized as $I(S,R)$. The next subsection presents the details of this maximization.

As we have been recently reminded, a model of Zipf's law for word frequencies should be able to make predictions beyond Zipf's law (Piantadosi 2014), and this is what applies to the family of optimization models above, which make successful predictions about the mapping of words into meanings (the principle of contrast), and vocabulary learning in children (Ferrer-i-Cancho 2017b). However, here we are going further: we are providing a set of general information theoretic principles, i.e. *a set of fundamental* (*but refutable*) *peculiar hypotheses* (as M. Bunge would put it), that can be used to build models in new domains, e.g., word order for the present article. Piantadosi's (2014) reminder falls short: the ultimate goal of a language researcher is not to design a model that predicts various properties of language simultaneously but to build a general theory for the class of linguistic phenomena.

## 2.1 The order that minimizes the uncertainty about the target or that maximizes its predictability

Suppose that a linguistic sequence (a sequence of words or a sequence of gestures) has $m$ elements. The sequence can be represented by $m$ random variables $X_1$, ...,$X_i$, ..., $X_m$, where $X_i$ represents some information about the $i$-the element of the sequence. The setup is abstract and thus flexible: $X_i$ could be the word type, the part-of-speech or the meaning of the $i$-th element of the sequence.

Suppose that the whole sequence consists of one target element and other $n = m - 1$ elements. For instance, the target element could be the head and the other elements could be the dependents (modifiers or complements). We use the random variable $Y$ for the target and $X_1,...,X_i,...,X_n$ for the other elements. Again, $Y$ could be the word type, the part-of-speech or the meaning of the target element of the sequence.

When $i$ elements have been produced,

- The uncertainty about the target $Y$ is defined as $H(Y|X_1, X_2,...,X_i)$, the conditional entropy of $Y$ given $X_1, X_2,...,X_i$.
- The predictability of the target is defined as $I(Y|X_1, X_2,...,X_i)$, the mutual information between $Y$ and $X_1, X_2,...,X_i$.

For instance,

- $H(Y|X_1, X_2,...,X_i)$ could be the uncertainty about the meaning of the target $Y$ (e.g., the predicate representing the meaning of the target according to logical semantics) when the speaker has produced the words forms $X_1, X_2,...,X_i$.
- $I(Y|X_1, X_2,...,X_i)$ could be the predictability of the meaning of the target $Y$ when the speaker has produced the word forms $X_1, X_2,...,X_i$.

We are interested in the placement of the target where its uncertainty is minimized or its predictability is maximized. Further mathematical details can be found in Appendix A. Here we explain the bulk of the arguments.

The problem of the optimal placement of the target can be formalized as follows. The solutions of

$$\text{argmin}_{1\leq i\leq n} H(Y|X_1, X_2,..., X_i) \quad (1)$$

yield the optimal placements according to uncertainty. For instance, if the solution was $i=n$ then the minimum would be reached when the target is placed last. If the solution was $i=0$ then minimum would be reached when the target is placed first. Similarly, the solutions of

$$\text{argmax}_{1\leq i\leq n} I(Y|X_1, X_2,..., X_i) \quad (2)$$

yield the optimal placements according to predictability. It can be shown that Eqs. 1 and 2 have at least one solution, i.e. $i=n$. Put differently, the optimal placement of the target is at least in the last position in a real linguistic sequence: real linguistic sequences exhibit long-range correlations both at the level of letters and at the level of words (Montemurro & Pury 2002, Ebeling & Pöschel 1994, Alvarez-Lacalle et al 2006, Moscoso del Prado Martín 2011, Altmann et al 2012). The argument relies on two crucial properties (Appendix A):

$$H(Y|X_1, X_2,..., X_{i-1}) \geq H(Y|X_1, X_2,..., X_i) \quad (3)$$

for $i \geq 1$, and

$$I(Y|X_1, X_2,..., X_{i-1}) \leq I(Y|X_1, X_2,..., X_i) \quad (4)$$

for $i \geq 2$. Equality in Eqs. 3 and 4 appears only in some particular cases (Appendix A).

The result in Eq. 3 and Eq. 4 allow one to understand why postponing the target (producing more elements of the sequence) is optimal. In general, the uncertainty about the target reduces as the target is postponed, and implies that the minimum uncertainty is reached when it appears last (Eq. 3). Similarly, the predictability of the target improves, in general, as the target is postponed and the maximum predictability will be reached at least when it is placed at the end of the sequence. Therefore, in the absence of further knowledge about a sequence, the optimal strategy is to put the target last.

To sum up, the minimization of the uncertainty of the target or the maximization of its predictability leads to a final placement of the target. Interestingly, the element that has to be put last depends on the target. For instance, if the target is the head then its dependents should appear first. In contrast, if the target is one of the dependents (e.g., the object of a verb) then the head should not appear last.

The argument can be refined considering the problem of the minimization of the energetic cost associated to the uncertainty or to predictability. In this case, we define two functions, i.e. $g_H$ and $g_I$, that translate, respectively, entropy and mutual information into an energetic cost from the perspective of uncertainty or predictability (thus these cost functions do not take into account dependency length minimization costs). In particular, $g_H$ is a strictly monotonically increasing function while $g_I$ is a strictly monotonically decreasing function.

Then the optimal placement according to uncertainty is given by

$$\text{argmin}_{1 \leq i \leq n} g_H[H(Y|X_1, X_2, \ldots, X_i)]. \tag{5}$$

while the optimal solution according to predictability is given by

$$\text{argmax}_{1 \leq i \leq n} g_I[I(Y|X_1, X_2, \ldots, X_i)]. \tag{6}$$

Again the optimal strategy in general is to put the target last in the absence of any further information.

$g_H$ and $g_I$ play the same role as the function $g$ that has been used to investigate the optimal placement of the head according to dependency length minimization (Ferrer-i-Cancho 2015a, Ferrer-i-Cancho 2014). In the latter case, $g$ is a strictly monotonically increasing function that translates an edge length into its energetic cost.

We have presented uncertainty minimization and predictability maximization as equivalent (Section 1). However, Eqs. 5 and 6 show that uncertainty minimization has a broader scope because $m \geq 2$ suffices to decide that the target should be placed last (when $m$ = 1 there is no decision to make). In contrast, predictability maximization needs $m \geq 3$. Therefore, uncertainty minimization can operate on smaller sequences than predictability maximization. Hereafter we will use uncertainty minimization by default bearing in mind that it is equivalent to predictability maximization when $m \geq 3$.

## 2.2 A conflict between uncertainty minimization and dependency length minimization

Suppose that a sequence consists of a head and $n = m - 1$ dependents. According to the principle of minimization of uncertainty, the optimal placement of the head is extreme: at the end if the target is the head or at the beginning if the target are the dependents seen as a block of consecutive elements (in the latter case, the dependents have to be placed last which implies that the head is placed first). In contrast, the optimal placement of the head is at the center according to the principle of dependency length minimization (Ferrer-i-Cancho 2015a), as illustrated in Fig. 2. If $m$ is even there only one central placement that is optimal. If $m$ is odd there are two central placements (Fig. 2).

This implies that these two order principles are in conflict provided that $m \geq 3$. To see that no conflict exists when $m < 3$ notice that no word order problem exists when $m < 2$. When $m$ is even, there are two central positions and if $m = 2$ any position is therefore optimal for dependency length minimization (Ferrer-i-Cancho 2015a). Therefore, one expects that word order is determined by uncertainty minimization when $m = 2$.

To understand the severity of the trade-off, notice that an extreme head placement (head first or head last), thus an optimal placement of the target according to uncertainty minimization, maximizes the cost of dependency lengths (Ferrer-i-Cancho 2015a). While an extreme placement of the head yields a maximum sum of dependency lengths that is (Ferrer-i-Cancho 2015a)

$$D = \binom{m}{2} = \frac{m(m-1)}{2}, \qquad (7)$$

a central placement of the heads gives a minimum sum of dependency lengths that is (Ferrer-i-Cancho 2015a)

$$D = \frac{1}{4}(m^2 - m \bmod 2). \qquad (8)$$

In sum, the best case for uncertainty minimization is the worst case for dependency length minimization.

Interestingly, the converse does not hold: the best case for dependency length minimization is not the worst case for uncertainty minimization. When the head is the target and it is placed at the center, it is preceded by some elements that may have helped to reduce its uncertainty. When the target is the dependents and the head is placed at the center, the head helps to reduce the uncertainty of the dependents that have not appeared yet.

# 3. Constant entropy rate and related hypotheses

## 3.1. An introduction

Here we compare our arguments about word order against the constant entropy rate (CER) and related hypotheses (Genzel & Charniak 2002, Levy & Jaeger 2007, Jaeger 2010), These hypotheses are argued to explain various linguistic phenomena, e.g., syntactic reduction (Levy & Jaeger 2007, Jaeger 2010) and the frequency of word orders (Maurits et al 2010). We review them here because they are considered as a reference theory to any alternative information theoretic approach to language by some language researchers. The importance of these hypotheses is evident from the number of citations, the impact factors of the journals, and the institutions from which they are broadcast.

The core of these hypotheses is the existence of a "*preference to distribute information uniformly across the linguistic signal*" (Jaeger 2010, p. 23). In greater detail, the hypothesis could be formulated as (Jaeger 2010, p. 24)

> "*Human language production could be organized to be efficient at all levels of linguistic processing in that speakers prefer to trade off redundancy and reduction. Put differently, speakers may be managing the amount of information per amount of linguistic signal (henceforth information density), so as to avoid peaks and troughs in information density.*"

## 3.2. The origins of the hypotheses

This idea was introduced by August and Gertraud Fenk (1980, 2nd paragraph from the bottom of page 402):

> "*A communication system, which is supposed to deliver messages without loss, should not only be required to have a certain average level of redundancy (not exceeding the short term memory capacity), but also, that the information is distributed as uniformly as possible across small time spans.*"[2]

and developed in a series of articles (see Fenck-Oczlon (2001) for a review). Figure 1 of Jaeger (2010) and the figure in p. 403 of Fenk & Fenk (1980) are similar in terms of the axes' names and the shape of the curves. The work by G. Fenk predates by about 30 years what are considered to be the core articles (Jaeger 2010, Jaeger & Levy 2007) and by about 20 years the foundational articles of this family of hypotheses (Genzel & Charniak 2002, Aylett & Turk 2004).

There is a general reference to Fenk-Oczlon (2001) in Jaeger (2010), detached from the context of uniform information density. The relevant passages of section "2.2 Frequency and the constant flow of linguistic information" of Fenk-Oczlon (2001) are not mentioned. In the following, we will use the label "constant flow hypothesis" to refer to the original formulation. The following sections are focused on the developments of the later hypotheses of Section 3.1.

## 3.3 Their formal definition and their real support

Constant entropy rate and related hypotheses are popular among cognitive scientists working on language. However, they are generally unknown to quantitative linguists and the physicists who started investigating the statistical properties of symbolic sequences in the 1990s (e.g., Ebeling & Pöschel 1994). This is not very surprising given the lack of contact between these different disciplines, but also given the large gulf that separates the formal statements of these hypotheses and the statistical properties of real language.

Suppose that $H(X_i|X_1, X_2,...,X_{i-1})$ is the entropy of $X_i$, the $i$-th type of the sequence, knowing the types that precede it. In mathematical detail, the constant entropy rate (CER) hypothesis states that $H(X_i|X_1, X_2,...,X_{i-1})$ should remain constant as $i$ increases, i.e. (Genzel, D. & Charniak 2002)

$$H(X_1) = H(X_2|X_1) = ... = H(X_i|X_1,...,X_{i-1}) = ... = H(X_m|X_1,...,X_{m-1}). \quad (9)$$

To a quantitative linguist familiar with Hilberg's law (Hilberg 1990), it is obvious that Eq. 9 does not hold since that law states that

$$H(X_i|X_1,...,X_{i-1}) \approx \mathrm{a}i^{-\gamma}, \quad (10)$$

where $\gamma \approx 0.5$ and $a$ is a positi.ve constant. A more plausible version of the law has been proposed, by Dębowski (2015), namely

$$H(X_i|X_1,...,X_{i-1}) \approx \mathrm{a}i^{-\gamma} + b, \quad (11)$$

where $a$ and $b$ are positive constants.

[2] We owe this translation from the original German version to Chris Bentz.

Therefore, real texts do not satisfy Eq. 9. However, Eq. 9 is satisfied when $X_1,...,X_i,...,X_m$ are independent identically distributed (i.i.d.) variables. Thus a text consistent with the constant entropy rate hypothesis is easy to generate: take a real text and scramble it at the desired level (e.g., letters or words). The random text that you will produce will fit CER beautifully at the level chosen.

Consider a concrete sequence $x_1,x_2,...,x_i,...,x_m$. A related hypothesis is the uniform information density (UID) hypothesis, that is defined on $p(x_i|x_1,...,x_{i-1})$, the probability of the *i*-th element of a sequence conditioned on the previous elements. The hypothesis states that (Levy & Jaeger 2007)

$$p(x_1) = p(x_2|x_1) = ... = p(x_i|x_1,...,x_{i-1}) = ... = p(x_m|x_1,...,x_{m-1}). \quad (12)$$

While testing the validity of CER is easy, as we have seen above, refuting UID is more difficult *a priori* because it is poorly specified. For this reason, more specific hypotheses have been defined from Eq. 12 (Ferrer-i-Cancho et al 2013a). The strong UID hypothesis states that Eq. 12 should hold in every sequence of length *m* that can be produced. The full UID hypothesis is a particular case of strong UID where the set of sequences that can be produced are all possible sequences (i.e. the Cartesian product of the sets of symbols available at every position). Strong UID is a particular case of CER and therefore both versions of UID suffer from all the limitations of CER. The full UID is a particular version of the strong UID that implies a sequence of independent elements.

A challenge for CER and UID is that they hold in situations that are incomepatible with language. A scrambled text satisfies CER, i.e. Eq. 9, with $H(X_i|X_1, X_2,...,X_{i-1}) = H(X)$ for $1 \leq i \leq m$, where $H(X)$ is the entropy of the words of the text. Other sequences also satisfy CER (Eq. 9) with $H(X_i|X_1, X_2,...,X_{i-1}) = 0$ for $1 \leq i \leq m$:

- A homogenous sequence, e.g., "*aaaaaa*…" (another example of a sequence of i.i.d. variables, notice $p(x_i|x_1,\ldots,x_{i-1}) = 1$ for $i \geq 1$ and for every $x_1,...,x_{i-1},x_i$ in the support set).
- A perfect periodic sequence, i.e. a sequence of that consists of the repetition of a block of $T$ different types, e.g.,"*abcabcabc*…". When $T = 1$ we have a homogenous sequence and thus the interesting case is $T > 1$. If we assume that $H(X_i|X_1, X_2,...,X_{i-1})$ is the entropy of the *i*-th element of the sequence given all the preceding elements then we have $H(X_i|X_1, X_2,...,X_{i-1}) = 0$ for $1 \leq i \leq m$ because the first element is always the same and the next element can always be predicted perfectly knowing the last element. If we relax the definition of $H(X_i|X_1, X_2,...,X_{i-1})$ as the entropy of the *i*-th element of an arbitrary subsequence of the original sequence given the preceding elements in that subsequence then we have quasi CER, namely $H(X_i|X_1, X_2,...,X_{i-1}) = \log T$ for i = 1 and $H(X_i|X_1, X_2,...,X_{i-1}) = 0$ for $2 \leq i \leq m$. The reasons is that the first element is one of the block chosen uniformly at random and the next element can still be predicted perfectly knowing the last element. A perfect periodic sequence with T > 1 shows that CER does not imply independence between elements.

Notice that a scrambled text and a homogeneous sequence are examples of sequences of independent and identically distributed (i.i.d.) elements. CER holds for any i.i.d. process but is not limited to them as the example of a perfect periodic sequence with $T > 1$ indicates.

Therefore, CER is satisfied by sequences that include the best case (a perfect periodic sequence) and the worst case (a sequence of identically distributed elements) for predicting the next element of the sequence. As a principle of word order, CER includes sequences that lack any order.

## 3.4 The justification of the hypotheses

The main argument used to justify the uniform information density and related hypotheses is the phenomenon of reduction, namely "*more predictable instances of the same word are on average produced with shorter duration and with less phonological and phonetic detail*" (see Jaeger 2010, p.23 for a review of the literature on this phenomenon). This context-dependent reduction is reminiscent of the tendency of more frequent words to be reduced regardless of their context (Fenk-Oczlon 2001). We will refer to the latter as 1st order reduction and to the former as higher order reduction.

Standard information theory is concerned about 1st order reduction. Suppose that $p_i$ and $l_i$ are, respectively, the probability and the length of the code of the $i$-th type and then

$$\sum_i p_i = 1. \tag{13}$$

Within coding theory, the goal of solving the problem of compression is to minimize the mean length of the codes assigned to each type (Cover & Thomas 2006, p. 110), i.e.

$$L = \sum_i p_i l_i, \tag{14}$$

under a certain coding scheme (typically uniquely decipherable codes). Put differently, coding theory is concerned about reducing the length of "words" as much as possible. Under the scheme of uniquely decipherable codes or non-singular codes, optimal coding successfully predicts Zipf's law of abbreviation, namely the tendency of more likely elements to be shorter (Ferrer-i-Cancho et al 2015). Therefore, standard information theory is concerned with reduction of more likely elements without context. Interestingly, standard information theory can be easily extended to reduction with context, namely higher order reduction. Suppose that we focus on the reduction of a concrete word $y$.

We may define the mean length of a type in combination with a previous context of $n$ consecutive words as

$$L_n = \sum_{x_1,x_2,\ldots,x_n,y} p(x_1,x_2,\ldots,x_n,y)\,l(x_1,x_2,\ldots,x_n,y), \tag{15}$$

where $p(x_1,x_2,\ldots,x_n,y)$ and $l(x_1,x_2,\ldots,x_n,y)$ are, respectively, the probability and the length of the type $y$ when it is preceded by the sequence of types $x_1,x_2,\ldots,x_n$. We assume

$$\sum_{x_1,x_2,\ldots,x_n,y} p(x_1,x_2,\ldots,x_n,y) = 1, \tag{16}$$

when $n = 0$, $L_n$ becomes $L$ as defined in Eq. 14. Again, optimal coding predicts a generalized Zipf's law of abbreviation: the tendency of more frequent type-context combinations to be shorter (Ferrer-i-Cancho et al 2015). A prediction under non-singular coding or uniquely decipherable encoding is that the minima of $L_n$ satisfy

$$\tau(p(x_1,x_2,\ldots,x_n,y), l(x_1,x_2,\ldots,x_n,y)) \leq 0, \tag{17}$$

where $\tau(\ldots,\ldots)$ is the Kendall tau correlation (Ferrer-i-Cancho et al 2015). This general result has strong implications for research on the reduction of a target type, e.g. "that" as in Levy and Jaeger (2007). In particular, a target type is expected to be shorter in contexts that are more likely. Put differently, compression predicts that types that appear in more predictable contexts have to be reduced.

To see it from a complementary perspective, we may define $L_n$ equivalently as

$$L_n = \sum_y L_n(y), \quad (18)$$

with

$$L_n(y) = \sum_{x_1, x_2, \ldots, x_n} p(x_1, x_2, \ldots, x_n, y) l(x_1, x_2, \ldots, x_n, y). \quad (19)$$

Renormalizing locally, i.e. dividing $L_n(y)$ by $p(y)$, we obtain

$$M_n(y) = \sum_{x_1, x_2, \ldots, x_n} p(x_1, x_2, \ldots, x_n | y) l(x_1, x_2, \ldots, x_n, y), \quad (20)$$

where $p(x_1,x_2,\ldots,x_n|y)$ is the probability of the block $x_1,x_2,\ldots,x_n$ knowing that it is followed by $y$.

Notice that

$$\sum_{x_1, x_2, \ldots, x_n} p(x_1, x_2, \ldots, x_n | y) = 1. \quad (21)$$

$M_n(y)$ can be seen as a particular case of $L$ where the set of types is defined by all the contexts of length $n$ that can precede a concrete type $y$. The minimization of $M_n(y)$ predicts that y should be shorter in more likely contexts (Ferrer-i-Cancho et al 2015). Again, a prediction is that the minima of $M_n$ (y) satisfy

$$\tau(p(x_1, x_2, \ldots, x_n | y), l(x_1, x_2, \ldots, x_n, y)) \leq 0. \quad (22)$$

Therefore, one does not need uniform information density and related hypotheses to explain reduction. The principle of compression can suffice.

A potential difference between first order compression and higher order compression could be that the latter may allow for types of length 0, namely full reduction, thanks to the preceding context or the function that the words that undergo total reduction perform. For instance, function words such as the conjunction "that" are easier to remove than content words. Such a tolerance to function word removal is the basis of telegraphic speech (Akmajian et al 2001, p. 23). In 1st order compression, non-singular coding implies codes of length greater than zero.

Interestingly, the case of full reduction and telegraphic speech could be regarded as cases of lossy compression. The critical question is: if lossless or lossy compression may account for reduction, why should CER or UID be necessary?

## 3.5. The link with standard information theory

A very important feature of a scientific field is that it must be

> "*a component of a wider cognitive filed, i.e. there is at least one other (contiguous) research field such that (a) the general outlooks, formal backgrounds, specific backgrounds, funds of knowledge, aims and methodics of the two fields have non-empty overlaps and (b) either the domain of one field is included in that of the other, or each member of the domain of one of them is a component of a system belonging to the other domain*" (Bunge 1984).

Research in the field of CER/UID and information theory overlap. The domain of CER/UID – human language – is a subset of the domain of information theory, that is also concerned with artificial systems as well as other means of information storage and transmission of information such as genomic sequences (e.g., Naranan & Balasubrahmanyan 2000) or animal behavior (e.g., McCowan et al 1999, Suzuki et al 2006). A very important component of a scientific theory is a formal background, namely "*a collection of up-to-date logical or mathematical theories (rather than being empty or formed by obsolete formal theories)*" (Bunge 1984). Followers of CER/UID employ jargon from standard information theory such as "*noisy channel*", "*channel capacity*" (e.g., Jaeger 2010, Piantadosi et al 2011), and posit strong links between information theory and uniform information density

> "*The hypothesis of Uniform Information Density links speakers' preferences at choice points during incremental language production to information theoretic theorems about efficient communication through a noisy channel with a limited bandwidth* (Shannon, 1948)" (Jaeger 2010, p. 25)

Does it mean that information theory is actually the formal background of CER/UID in a Bungean sense?

Mentions of standard information theory such as the ones given above could be neglected if CER/UID were not considered reference theories to alternative approaches based on information theory, such as ours. However, since they are widely considered as such it is worth scrutinizing in more detail their actual links with information theory. As we have shown above, followers of CER/UID fail to identify the phenomenon of reduction as a manifestation of compression, thus missing a link with standard coding theory. The loose connection with standard information theory can be understood further when revising the predictions of CER/UID on the efficiency of language.

One of the major problems of CER/UID and related hypotheses is that they are presented as arising from efficiency considerations, but the exact link with optimization is unclear. One example is an article that makes strong claims about the efficiency of language but does not specify the cost function that is being minimized (Piantadosi et al 2011). A complete argument about optimization requires at least three fundamental components:

1. A cost function
2. A theoretical insight linking the minimization of that function and statistical properties of the system.
3. A baseline

In standard coding theory, $L$ (Eq. 14) is the cost function. If $L$ is minimum then it is well-known that

$$l_i = \lceil -\log p_i \rceil \tag{23}$$

for uniquely decipherable encoding (Cover & Thomas, 2006). The three components are found in an extension of coding theory for research on Zipf's law of abbreviation in natural communication systems (Ferrer-i-Cancho et al 2013b, Ferrer-i-Cancho et al 2015): there the cost function is the generalization of $L$ and various mathematical arguments are used to show the relationship between Zipf's law of abbreviation and the minimization of that cost function. The baseline is defined by a randomization of the mapping of probabilities into lengths.

The theoretical insight is crucial. Without it, it is easy to make wrong inferences. Finding a strong correlation between a measure of "information content" and length does not imply that speakers are making optimal choices involving the contexts where words appear (Piantadosi et al 2011): a linear dependency between these two variables may simply arise internally, from the units making a word (e.g., letters) as random typing shows simply (Ferrer-i-Cancho & Moscoso del Prado Martín 2011). Paradoxically, destroying a text by scrambling the text sequence (at the level of words or at the level of characters) will produce a sequence of i.i.d. words that exhibits perfect agreement with CER. Furthermore, finding a correlation between "information content" and word length that is stronger than the correlation between frequency and length of Zipf's law of abbreviation (Piantadosi et al 2011) does not imply that the former correlation is the outcome of a higher degree optimization: in case of optimal coding, a perfect correlation between frequency and length is not expected due to ties of length in optimal codes (Ferrer-i-Cancho et al 2015). For instance, Eq. 23 implies that all types with the same probability should have the same length and frequency ties are many in real texts (frequency ties are beautifully described by Zipf's number-frequency law, Zipf 1949). The lack of a cost function and a theoretical understanding of its predictions can lead to wrong inferences.

As far as we know, the only attempt to derive mathematically uniform information density from cost minimization can be found in Levy & Jaeger 2007. The attempt is partial for two reasons: it depends on a parameter $k$ and cost is only minimized for $k < 1$. The value or the range of values of $k$ that are suitable for human language are unclear. Second, it does not concern CER, which is a more general condition than strong UID (Section 3.3).

## 3.6. CER and UID versus our word theory order

It is important to notice that both CER and our entropy minimization principle for word order are hypotheses on conditional entropies. However, there are some differences between our word order theory and CER/UID that are worth reviewing:

1. While in CER the target of conditioning is moving (Eq. 9), in our case the target is constant (Eq. 1).
2. While CER applies even to sequences that lack any order (namely to sequences of independent and identically distributed elements), our approach relies heavily on statistical dependencies among elements of the sequence (in sequence of independent elements, postponing the target will not help to predict it).
3. While the major statement of CER is a hypothesis which real language does not satisfy, our hypothesis is based on a basic truth, that "*conditioning reduces entropy*" in general, and this predicts the optimal placement of costly elements. The latter is not an opinion, conjecture or a hypothesis, but a mathematical fact. The same applies to the optimal placement according to dependency length minimization and the conflict between uncertainty minimization and dependency length minimization. Notice that the original "constant flow hypothesis" was also based on the fact that "conditioning reduces entropy" (recall the quote of Fenk-Oczlon (1989) in the introduction).
4. While CER and UID are presented as primary overarching principles, a conflict between principles is at the core of our theoretical approach. CER and UID are concerned about the trade-off "between redundancy and reduction" (Jaeger 2010), but only in the periphery of the argument. In contrast, the core of our theory defines word order as multiconstraint satisfaction problem where a principle of entropy minimization is in conflict with the principle of dependency length minimization (Section 2). Because of the secondary importance of distorting factor and conflicts between principles in CER and UID (Jaeger 2010, Levy & Jaeger 2007), these hypotheses are seen as incomplete (Ferrer-i-Cancho et al 2013a). Classic examples of linguistic theory where conflicts are at the core are G. K. Zipf's, whose view is based on conflicts between hearer and speaker needs (Zipf 1949), as well as R. Köhler's synergetics (Köhler, 1987; Köhler, 2005). A spin-off of Zipf's view are model of Zipf's law for word frequencies that are based on the conflict between the minimization of the entropy of a vocabulary and the maximization of the mutual information between words and meanings (Ferrer-i-Cancho 2005, Ferrer-i-Cancho & Solé 2003).

5. UID and related hypotheses are concerned with a trade-off "between redundancy and reduction" (Jaeger 2010) that are symmetric terms: one is simply the opposite of the other. In contrast, our theory is concerned with a trade-off between two non-symmetric principles: uncertainty minimization and dependency length minimization. In standard information theory, there are no trade-offs between redundancy and reduction *per se* but trade-offs between different goals. In the terminology of information theory, goals define problems (Cover & Thomas 2006). Roughly speaking, the solution to the problem of transmission leads to increased redundancy while the solution to the problem of compression leads to reduction (Cover & Thomas 2006).
6. While CER and related hypotheses appear disconnected from optimization models of communication (Ferrer-i-Cancho 2017a), our approach to word order extends the domain of application of two fundamental principles of these models, i.e. entropy minimization and mutual information maximization. These optimization principles are relevant for their capacity to shed light on the origins of Zipf's law for word frequencies, the principle of contrast and a vocabulary learning (Ferrer-i-Cancho 2014).
7. While the connection between standard information theory and the UID/CER hypotheses is weak, various connections are already available between optimization models of communication and information theory through the problem of compression or model selection (Ferrer-i-Cancho 2017a).
8. CER and related hypotheses suffer from a psychological bias: they stem from a view where linguistic phenomena (word order in particular) are caused by absolute constraints of the human brain. In contrast, our framework is open to other causes: certain word order features may simple increase the survival over time of dominant word orders (Ferrer-i-Cancho 2015a). These other causes my exploit constraints of the human brain and then these constraints would not be the ultimate reason for the observed phenomena.

Points 5-7 are very important because a scientific theory should be more than a collection of disconnected ideas, as Bunge (2013) reminds us.

### 3.7. Ways to improve CER

Proponents of these hypotheses may argue that the disagreement with Hilberg's law does not reject their hypothesis because their true definition is that languages should tend towards to CER or UID (whether they actually reach Eq. 9 or Eq. 12 is irrelevant or secondary). However, such a disagreement implies two fundamental questions:

1. Why should languages tend towards CER or UID (Eq. 9 or Eq. 12)?
2. Why are languages not reaching CER or UID?

As we have seen in Section 3.5, the answer to Question 1 is unclear because a sufficiently developed theory is not available: the cost function that should accompany any claim on optimization and other fundamental components of a real theory are missing in general. Such a theoretical understanding is also lacking for Question 2. With such incompleteness and under -specification (to the extent of being fully satisfied by i.i.d. processes, Section 3.3), it is rather straightforward to fit these hypotheses to a wide range of phenomena. Would the explanatory power of these theories remain constant if they were specified in greater detail? More importantly, are these hypotheses really necessary? We have argued that compression is an alternative hypothesis with higher predictive power (Section 3.4). A further challenge for their need will be provided below.

The disagreement between CER and Hilberg's law forces one to see CER as a tendency and this has the drawback of reducing the precision of the hypothesis. The constant entropy rate hypothesis can be relaxed with precision in a way that does not contradict that law. The goal is to avoid peaks of information by reducing the conditional entropy from the very beginning (Fenk-Oczlon 1989). The problem can be formalized as the minimization of the following cost function:

$$\max_{x_i} H(X_i|X_1, X_2, \ldots, X_{i-1}). \tag{24}$$

This is equivalent to minimizing $H(X_1)$ in a real linguistic sequence thanks to Hilberg's law (Appendix B). Put differently, peaks could be reduced simply with a bias to minimize the entropy of the initial elements assuming Hilberg's law.

Then, we do not need to invent a new principle: the minimization $H(X_1)$ can be seen as an example of a general principle of entropy minimization that has been applied to shed light on the origins of Zipf's law and that could be an indirect consequence of the minimization of $L$, namely compression (Ferrer-i-Cancho 2017a). Therefore, our attempt to improve CER adds another reason to not need CER (recall Section 4.3).

In sum, there is no objective reason to regard CER and related hypotheses as reference theories.

## 4. Discussion

### 4.1 The word order predicted by minimizing uncertainty

Section 2 provides a general argument for the placement of a target element of the sequence: it should be placed last to minimize its uncertainty. The argument is general but under-specified till we choose a target. We may choose the target between the head and its dependents or between the predicate and its arguments.

The word order problem has two symmetric solutions depending on the target:

- If the target is the head, uncertainty minimization predicts that the target should be placed last.
- If the targets are the non-head elements (the arguments), the prediction is that the non-head elements should be placed last, which implies that the head should be placed first.

Therefore our findings have implications for branching direction theory (Dyer 2011): left-branching minimizes the uncertainty of the head and right-branching minimizes the uncertainty of the dependents (complements/modifiers). Notice dependency length minimize-ation can also produce consistent branching once the main verb has an extreme placement (Ferrer-i-Cancho 2008, Ferrer-i-Cancho 2015a).

Our general argument is that the most costly element should be the target, allowing one to break the initial symmetry between targets. Some costs that may determine the choice of the target will be presented below.

Let us consider the particular case of the ordering of the triple defined by S (subject), V (verb) and O (object). The following discussions assume that the verb is the head and is also valid for their semantic correlates: actor, action and patient (e.g., Goldin-Meadow et al 200; Langus and Nespor 2010). First, we consider that the target is either the head or the dependents. This yields that

- SOV and OSV are optimal when the target is the verbal constituent.
- VSO, VOS are optimal when the target is the non-verbal constituent.

The statistics of word orders suggest that verbs and their arguments are not symmetric targets. 89% of world languages that show a dominant word order do not put the verb first (Table 1). The *a priori* symmetry between verb initial and verb final languages can be broken in favour of verb final languages taking into account that verbs are harder to learn than nouns (Saxton 2010), which are the heads of the verbal complements (subject and object). For children, nouns are easier to learn than verbs (e.g., Imai et al 2008, Casas et al 2016), and actions (typically represented by verbs) are harder to pick up, encode and recall than objects (typically represented by nouns) (e.g., Gentner 1982, Gentner 2006, Imai et al 2005). Verb meanings are more difficult to extend than those of nouns (e.g., Imai et al 2005). Also, see McDonough et al 2011 for an overview of arguments on the difficulty of verbs as compared to nouns. Furthermore, arguments for the greater difficulty of verbs for infants can easily be extended to adults beyond the domain of learning. For these reasons, a communication system that aims at facilitating the processing and the learning of the most difficult items, i.e. verbs, may favour the strategy of minimizing the uncertainty about the verb (leading to verb last) over the strategy of minimizing the uncertainty about the nouns (leading to head first). The suitability of a verb last placement is supported by computer and eye-tracking experiments which indicate that the arguments that precede the verb help to predict it (Konieczny & Döring 2003).

The argument can be refined splitting dependents (arguments) into subjects and objects. By considering each of the elements as the target we get the optimal orderings (orderings that either minimize the uncertainty about the target or maximize the predictability about the target):

- The orders SOV and OSV are the optimal when the verb is the target.
- The orders SVO and VSO are the optimal when the object is the target.
- The orders VOS and OVS are the optimal when the subject is the target.

Again, the symmetry can be broken taking into account that verbs are harder to learn. In this case, SOV or OSV are expected. Interestingly, SOV is a verb final order that

- Covers 43.3 % of dominant orders in languages (65.3% in families) according to Table 1.
- Is hypothesized to prevail in early stages of evolution of spoken (Gell-Mann & Ruhlen 2011, Newmeyer 2000, Givon 1979) and signed languages (Sandler 2005, Fisher 1975).

- Is recovered in experiments of gestural communication (Goldin-Meadow et al 2008, Langus & Nespor 2010).
- Appears in *in silico* experiments under pressure to maximize predictability (Reali & Christiansen 2009).

The problem is that our optimality argument also predicts OSV, that covers only 0.4% of dominant orders in languages (0.3% in families) according to Table 1.

Our argument predicting verb final placement can be refined to yield only SOV in four ways:

- Assuming a hierarchy of multiple targets, namely the verb is the main target and the object is a secondary target. That would give the subject is placed first for not being a target and that the verb is put last for being the main target. Then SOV would follow. The idea is reminiscent of the standard approach to word order in typology that consists of assuming pairwise word order preferences (Cysouw 2008).
- Postulating an agent or subject bias that determines that the subject is placed first, the so-called agent first pragmatic rule (Schouwstra & de Swart 2014).
- As more frequent elements are put first (due to some psychological preference), the subject would be put first (Fenk-Oczlon, 1989). The argument is interesting for connecting the frequency effects that are used to justify the minimization of entropy in optimization models of communication (Ferrer-i-Cancho 2017a) with word order and thus this option has the potential to yield a compact theory of language with respect to the two preceding alternatives.
- An indirect effect of a hidden attraction towards SVO (see Section 4.6).

These possibilities should be the subject of further research.

Let us move to the problem of the optimal placement of dependents within the nominal constituents. For simplicity, we consider that the target are the head or the dependents. This yields that

- Placing the dependents before the nominal head is optimal when the target is the nominal head.
- Placing the dependents after the nominal head is optimal when the target are the dependents.

Thus the principle of uncertainty minimization could contribute to explain why no language consistently splits its noun phrases around a central nominal pivot, with half of the modifiers to the left and half to the right, as expected from the principle of dependency length minimization (Ferrer-i-Cancho 2015a). Support for this possibility comes from the complex interaction between dependency length minimization and other factor at short ranges (Gulordava et al 2015). However, uncertainty minimization does not need to be the only reason for this phenomenon: we have argued that the actual placement of modifiers could be the result of competition between dominant orders struggleing for survival (Ferrer-i-Cancho 2015a). However, believing that predictability maximization is the only reason why dependents of nominal heads tend to be put at one side of their head is theoretically naïve, because the principle of dependency length minimizeation at global scale predicts that those dependents are placed before the nominal head in verb final languages and after the nominal head in verb initial languages (Ferrer-i-Cancho 2008; Ferrer-i-Cancho 2015a). Therefore, dependency length minimization and uncertainty minimization can collaborate to yield an asymmetric placement of dependents at short ranges and may explain the origins of consistent branching in languages.

In Section 2, we have provided some mathematical results to understand the placement of heads in single head structures. More realistic scenarios should be investigated. Our theoretical framework should be extended to the case of the multiple head structures that are found in of complex sentences.

## 4.2 The optimality of word orders

Integrating the arguments of Section 4.1 with the predictions of dependency length minimization one obtains the following optimality map:

- SOV and OSV are optimal according to the minimization of the uncertainty about the verb, 43.7% of dominant orders in languages (65.6% in families) according to Table 1.
- SVO and OVS are optimal according to the minimization of dependency lengths, 41.1% of dominant orders in languages (15.8% in families) according to Table 1.
- VSO, VOS are optimal according to the minimization of the uncertainty about the non-verbal constituents, 13.9% of dominant orders in languages (15.9% in families) according to Table 1.

More precise optimality arguments can be built splitting the non-verbal constituents into subjects and objects or assuming a hierarchy or targets as explained in Section 4.1.

Our findings on the optimality of word orders and on the properties of the adaptive landscapes that optimality principles define (Section 2.3) are particularly relevant for researchers who have "no evidence that SOV, SVO, or any other word order confers any selective advantage in evolution" (Gell-Mann & Ruhlen 2011). Interpreting the diversity of word orders (Table 1) or the rather large proportion of languages lacking a dominant (about 2.4% of languages according to Table 1 but 13.7% according to Dryer 2013) as an absence of principles or adaptive value is theoretically naïve: it may simply reflect the difficulty for complying with incompatible constraints (Ferrer-i-Cancho 2014). The diversity of word orders would not be a manifestation of arbitrariness but an inevitable consequence of a multiconstraint optimization problem where the availability of word orders is constrained by a word order permutation ring.

## 4.3. Word order conflicts

The simple optimality map presented above clearly shows that there are at least two conflicts between principles: one internal to uncertainty minimization, i.e. the optimal order depends on the target of uncertainty minimization and another external, between dependency length minimization and uncertainty minimization.

The external conflict is due to the fact that the principle of dependency length minimization predicts that the head should be placed at the center while the principle of uncertainty minimization predicts that it should be placed at one of the ends (Section 2).

It is worth considering the interplay between dependency length and uncertainty minimization with the head as the target as the head moves from the beginning of the utterance to the end. Postponing the head minimizes its uncertainty but the cost of dependency lengths will depend on its placement (Ferrrer-i-Cancho 2015a). The cost of dependency lengths decreases as the head is postponed before the center of the sequence. From then on, dependency length costs will increase as the head is postponed. Put in technical terms, the landscape of dependency lengths as a function of the position of the head is quasi-convex for the case of a single head (Ferrer-i-Cancho 2015a). Put differently, dependency length minimization and postponing the head to minimize its uncertainty are 'allies' during the first half of the sequence and 'enemies' in the second half. In contrast, dependency length minimization and bringing the head forward to minimize the uncertainty about dependents are 'enemies' in the first half of the sequences and 'allies' in the second half.

Notice that the external conflict above arises in the context of the optimal placement of one head and its dependents. The problem is more complex if one considers further levels of organization: e.g., from the head verb to the heads of its complements and then from the heads of these complements to their dependents. In Section 4.1, we have shown that the principles can be in conflict at the level of the optimal placement of the verb but collaborate at the level of the placement of dependents of nominal heads.

G. Heyer and A. Mehler (2009) made us notice that the conflict between predictability (uncertainty) and dependency length minimization could be seen as a conflict between long term memory, that stores the probabilistic information underlying the definition of uncertainty or predictability, and online memory, where pressure to minimize dependency lengths originates. To Heyer & Mehler, conflicts between principles are reminiscent of conflicts between time cost and memory cost in algorithmic theory (Cormen et al 2009).

Since we have argued that any word order can be optimal *a priori* (e.g., any placement of V with respect to S and O is optimal for some reason), and that different orders are in conflict (e.g., putting the V at one of the two ends is against dependency length minimization), it is tempting to conclude that "*anything goes*" (any word order is valid). However, this is not our view. We believe that word order is determined at least by the experimental or ecological conditions (and previous history as we will see below). Examples of ecological conditions are the proportion of L2 speakers and the proportion of deaf individuals of the community. Examples of experimental conditions are the length the sequences to be uttered or gestured or the amount of pressure to maximize predictability as we will see immediately.

If additional pressure for predictability maximization is added *in silico,* experiments show that a verb final language (SOV) emerges, as expected from the theoretical arguments above (see Ferrer-i-Cancho 2014 for further details about this experiment). In general, verb final or verb initial languages are more likely in simpler sequences while verb medial languages are expected in more complex sequences (Ferrer-i-Cancho 2014). The case of verb initial languages could be special because they originate from verb medial languages (Gell-Mann & Ruhlen 2011) and therefore their sequential complexity does not need to be as low as that of the verb final languages that are typically found at early stages of word order evolution. Finally, recall that on top of theoretical arguments indicating that any word order can be optimal for some reason, we have added a further factor that could break the symmetry between word orders, such as the higher intrinsic difficulty of certain words, that would increase the chance that they are chosen as targets, or the recency or frequency effects, by which subjects would be put first (Section 4.1).

The explanatory power and potential of word order conflicts is illustrated by their capacity to shed light on the origins of word order diversity, on the phenomenon of languages lacking a dominant word order, on word order reversions in historical developments, and on alternative orders with a verb at the center (Ferrer-i-Cancho 2014). Furthermore, they also allow one to understand why real sentences do not achieve the minimum sum of dependency lengths that is expected if dependency length minimization was the only principle (Ferrer-i-Cancho 2004). Subsection 4.4 provides an updated account on word order diversity.

## 4.4. Word order diversity

Word order diversity can be interpreted in two ways: externally, comparing the variation of dominant word orders across languages, and internally, looking at the word orders that are adopted within a language.

Concerning external diversity, the optimality map presented in Section 4.2 shows that all verbal placement are optimal for some reason *a priori*. Adding that word orders are in conflict one expects that there is no single winner. Indeed, the six possible orders of subject, verb and object are found in languages (Table 1). We believe that the principles of word order and their conflicts have the potential to explain word order diversity (in combination with other components such as the word order permutation ring that will be reviewed later on). Although all verbal placements are optimal according to some word order principle, word order could be biased towards verb medial due to the increasing pressure for dependency length minimization as linguistic complexity increases (Ferrer-i-Cancho 2014), and also towards verb final due the higher complexity of verbs (McDonough 2011, Gentner 2006). We do not mean that the counts of word orders are unequivocally indicative of the degree of optimality of a word order because of word order evolution. A full explanation of the diversity of dominant word orders requires acknowledgement of the fact that word order evolution is a *path dependent process* where the initial word order is critical. Section 4.4 sheds some light on how the bulk of word order diversity can be generated, step by step.

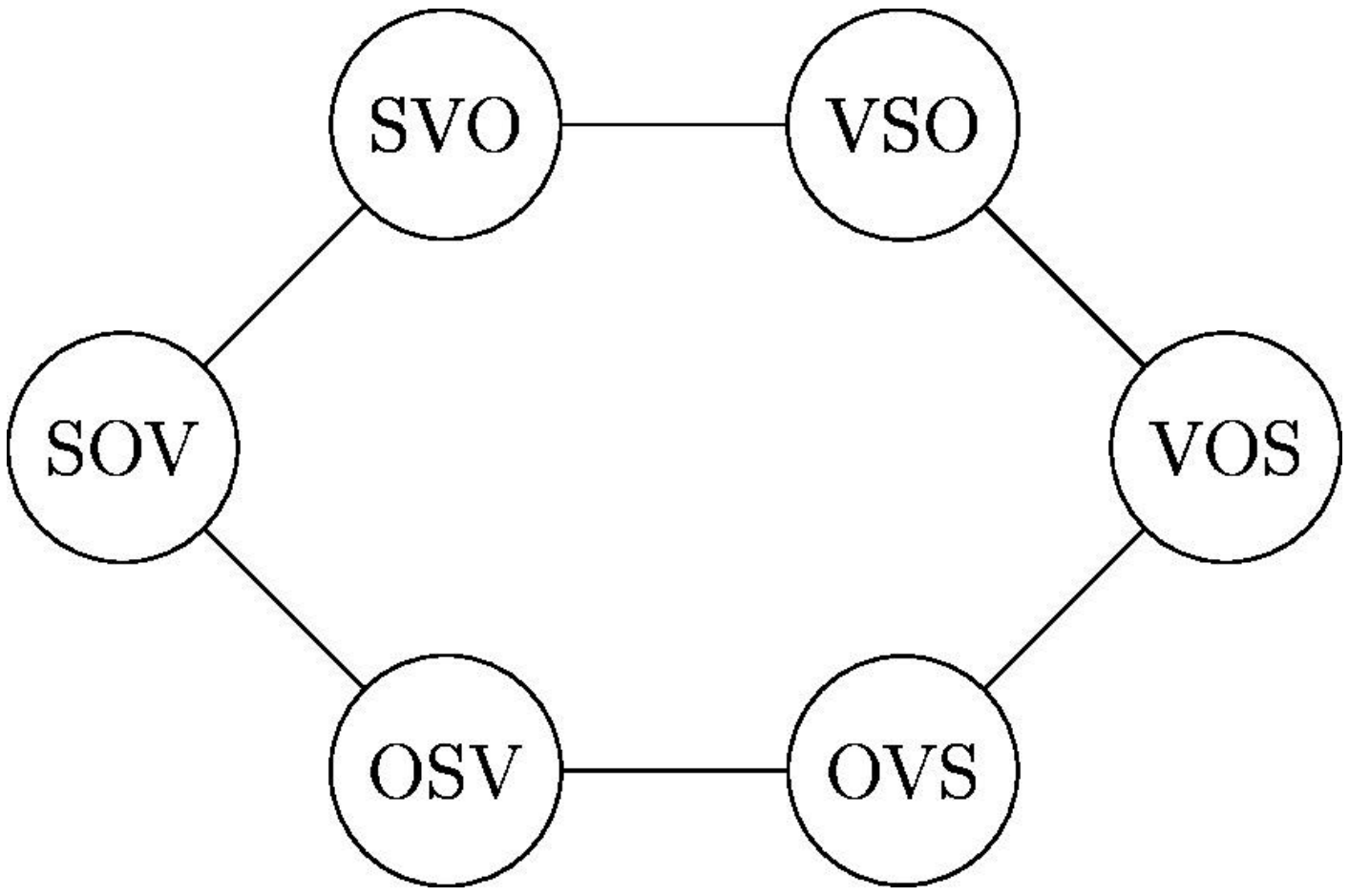

Figure 3. The permutation ring defined by all the 6 possible orderings of subject (S), verb (V), and object (O). Two orderings are connected if one leads the other after swapping two adjacent elements.

Internal word order diversity has been hypothesized to be constrained by a word order permutation ring that determines how a new word order can be generated from another (Ferrer-i-Cancho 2016). The *a priori* probability of a variant is hypothesized to be a monotonically decreasing function of the distance between the variant of the dominant order in a permutation ring (Fig. 3). The word order permutation ring beats the standard model of typology in explaining the composition of the couples of primary alternating word orders (Ferrer-i-Cancho 2016).

The power of the permutation ring to explain the evolution of the dominant word order will be reviewed in Section 4.5.

## 4.5 Word order evolution

Here we revisit the framework for word order evolution that has been presented in a series of articles for the evolution of the dominant ordering of subject, verb and object in languages (Ferrer-i-Cancho 2008, Ferrer-i-Cancho 2014, Ferrer-i-Cancho 2015a, Ferrer-i-Cancho 2016a). This framework has two major components: an early or initial order and transitions between orders.

Converging evidence supports SOV (or its semantic correlate actor, patient, action) as an initial or early stage in evolution (Gell-Mann & Ruhlen 2011, Langus and Nespor 2010, Pagel 2009, Goldin-Meadow et al 2008, Sandler et al 2005, Newmeyer 2000, Givon 1979, Fisher 1975). The early or initial word order is determined by conditions that facilitate the dominance of maximization about the predictability of the verb over either (1) dependency length minimization or (2) the minimization of the uncertainty about the other components. We have argued that the victory of the maximization of the predictability of the verb is likely to be determined by a series of factors:

- *The length of the sequences*. At early stages, linguistic sequences (of words or gestures) are expected to be shorter (Ferrer-i-Cancho 2014). This is easy to see in the extreme case of sequences of length two: the placement of the head is irrelevant for dependency length minimization but to minimize the uncertainty about the head, the verb should appear last. The size of the sequence where dependency length minimization can be neglected may be determined by the capacity of short term memory, i.e. about four elements (Cowan 2000).
- *Morphology*. Case marking facilitates the processing of SOV structures (Lupyan & Christiansen 2002).

In section 4.1 we have provide arguments for a preference for SOV over OSV.

Transitions are hypothesized to be constrained by the structure of the space of possible transitions and conditions that help one principle to dominate in the struggle between dependency length minimization and uncertainty minimization (or predictability maximize-ation). The space of possible transitions has been hypothesized to be determined by the minimum number of swaps of adjacent constituents that are needed to reach a word order from the current word order (Ferrer-i-Cancho 2008, 2015, 2016a). The *a priori* probability of a transition is hypothesized to be a monotonically decreasing function of the distance between the source order and the destination order in a permutation ring (Fig. 3). The transition from SOV to SVO is more likely *a priori* than the transition from SOV to OVS (the former requires only one swap; the latter requires two swaps). This is known as the *word order permutation ring hypothesis*. Further conditions may operate on this permutation ring, possibly distorting the predictions that can be made if the ring was the only constraint.

We will use the main path for word order evolution namely the transition from SOV to SVO and the transition from SVO to VSO/VOS to illustrate how these conditions apply (Gell-Mann & Ruhlen 2011). A striking feature of these transitions is that they involve source and destination orders that are adjacent or almost adjacent in the word order permutation ring (Fig. 3).

Table 3
Predictions on the most likely transition from SOV. Yes and No indicate presence or absence of the feature indicated in header of the corresponding column.

| Word order permutation ring | Dependency length minimization | Most likely destination |
|---|---|---|
| Yes | No | SVO and OSV |
| No | Yes | SVO and OVS |
| Yes | Yes | SVO |

According to the permutation ring, the most likely transitions from SOV are SVO and OSV. However, the typical destination from SOV is SVO (Gell-Mann & Ruhlen 2011). We hypothesize that the tie is broken in favour of SVO by the principle of dependency length minimization that predicts that the head is placed at the center and factors that may favour SOV over OSV exposed in Section 4.1. However, this opens a new problem since there are two orders with the verb at the center, i.e. SVO and OVS. Interestingly, OVS is farther away from SOV in the word order permutation ring. Thus, we conclude that SVO is the most likely transition. A summary of the argument is provided in Table 3.

Since we have argued above that a main raison for SOV to be the initial or early stage is the victory of the minimization of the uncertainty about the head over other principles, it is reasonable to think that SOV will be abandoned when the sequence complexity (sequence length) increases. That increase facilitates the victory of dependency length minimization (Ferrer-i-Cancho 2014). The chances of success of the transition increase under further conditions that prevent regression to SOV:

- SVO languages that put adjectives after the noun are more likely to stabilize because this relative placement of adjectives is neutral for SVO but inconvenient for SOV from the perspective of dependency length minimization (Ferrer-i-Cancho 2015). Interestingly, the number of SVO language with that peculiar placement of adjectives is above chance.
- Case marking facilitates the learning of SOV structures (Lupyan & Christiansen, 2002). The need of case marking for a more efficient processing of SOV is supported by Greenberg's universal 41, stating that SOV languages almost always have case marking (Greenberg 1963). Thus, regression to SOV could be harder from SVO languages lacking case marking. In turn, as languages with a high proportion of L2 speakers tend to lose case marking (Bentz & Winter 2013), the proportion of L2 speakers is likely to be one of the factors that determines the stabilization of a dominant SVO order, expanding the predictions of the Linguistic Niche Hypothesis (Dale & Lupyan 2012) to the domain of word order.

Once a language is SVO why should it become VSO/VOS? Again the permutation ring and certain conditions can explain the transition. Once a system has reached SVO the permutation ring offers two main possibilities: to come back to SOV or to move forward towards VSO (or VOS with less probability). Adding pressure to minimize the uncertainty about the nominal heads then VSO appears as the most like solution. VOS is among the second best solution for being only one step farther in the permutation ring with respect to SVO and putting the verb first optimally as VSO. A summary of the argument is provided in Table 4.

Table 4
Predictions on the most likely transition from SVO. Yes and No indicate presence or absence of the feature indicated in header of the corresponding column

| Word order permutation ring | Minimization of the uncertainty about the nominal constituents | Most likely destination |
|---|---|---|
| Yes | No | SOV and VSO |
| No | Yes | VSO and VOS |
| Yes | Yes | VSO |

The likelihood of the transition to VSO/VOS is increased by adaptations in SVO that prevent regression to SOV that preadapt SVO for VSO/VOS: placing adjectives after the nominal head is convenient for VSO/VOS (Ferrer-i-Cancho 2015a) but not for SOV.

The scenario of word evolution presented above strongly suggests that word order evolution is a *path dependent process* (Ferrer-i-Cancho 2016a, Ferrer-i-Cancho 2015a, Dunn et al 2011).

It is worth noting that the number of languages (or the number of families) with a certain dominant word order decreases as one moves in the permutation ring in a clock-wise sense (Table 1, Fig. 3). It has been argued that word order evolution may not have reached a steady state or equilibrium (Gell-Mann & Ruhlen 2011). With our arguments above, we do not mean that SOV or SVO may not exist anymore, as dominant word orders, in the future (if no new languages were created). From our arguments above it follows that the dominance of dependency length minimization (SVO) is easier to achieve but harder to abandon, because of the time elapsed since the birth of these languages and the ecological conditions of many linguistic communities. As for the former, the length and the complexity of sentences has probably increased over time (it is unlikely that in the birth of a language from scratch long sentences are used). The adoption of a writing system or access to higher education are relevant ecological variables for this growth as they facilitate the creation of longer and more complex sentences where dependency length minimization is critical. If these conditions remain, there is no reason to believe that in the future languages will tend to go back to extremely short sentences where SOV is easier to handle. Due to this fundamental pressure for dependency length minimization and the importance of the verb as a target, transitions beyond SVO (and SOV) could be secondary.

## 4.6. Word order diversity in the light of evolution

The fact that SOV and SVO cover the overwhelming majority of dominant orders in languages (Table 1) could result from a tree-fold combination

1. The initial preference for SOV a word permutation ring constraining possible moves.
2. A bias to reduce the uncertainty of the head.
3. Dependency length minimization.

In this view, the initial preference for SOV and the word permutation ring are crucial to understand the evolutionary history of word order as a path dependent process (Ferrer-i-Cancho 2016a, Ferrer-i-Cancho 2015a, Dunn et al 2011). Following the three-fold hypothesis, we revise the frequency of six possible orders classifying them according to the position of the verb:

- SVO, OVS (central verb) is a compromise between dependency length minimizeation and the minimization of the uncertainty of the head: the placement of the verb at the center is optimal according to dependency length minimization and in-between its best placement (last) and its worst placement (first) according to minimization of the uncertainty of the head (Section 2.2). The low frequency of OVS could be explained by the evolutionary history.
- SOV, OSV (verb last) satisfies the optimality of the principle of postponing the head. The low frequency of OSV could be explained by three facts 1): the initial preference for SOV 2) an attraction towards SVO due to pressure for dependency length minimization 3) OSV is farther from SVO than SOV according to the permutation ring.
- VSO/VOS (verb first) should have lower frequency because placing the verb first is the worst case for both principles. Additionally, they might be under represented due to the evolutionary history.

## 4.7. A general theory of word order and beyond

In this article, we have made one step forward to building a coherent theory of word order. The major components of the theory are

- A subtheory of word order from the dimension of dependency length minimization (Ferrer-i-Cancho 2008, 2015a,b).
- A subtheory of word order from the dimension of uncertainty minimization or predictability maximization (this article).
- An integrated subtheory of word order that explains how these principles interact: their conflict and the factors that determine the dominance of one over the other (this article and Ferrer-i-Cancho 2014).
- A subtheory of word order variation, both internal, i.e. within a language (Ferrer-i-Cancho 2016a) and also externally, i.e. across languages (this article and Ferrer-i-Cancho 2014).
- A subtheory of word order evolution (Ferrer-i-Cancho 2014, Ferrer-i-Cancho 2016a).

These subtheories are not collections of disconnected ideas. The subtheory of word order evolution relies on the assumption that word order evolution operates on constraints on word order variation (Ferrer-i-Cancho 2016a). These subtheories are unified through the word order permutation ring. In turn, this ring and the principle of dependency length minimization stem from a general principle of distance minimization (Ferrer-i-Cancho 2016a). Beyond word order the theory is connected with the theory of Zipf's law for word frequencies: both the minimization of uncertainty and predictability maximization follow from a general principle of entropy minimization and mutual information maximization that can be applied to shed light on the origins of Zipf's law for word frequencies.

The theory is articulated by key traversal concepts:

- *Intrinsic conflicts*: word order principles are intrinsically in conflict as we have seen in this article. These conflicts may underlie the diversity of dominant word orders found across languages as well as the lack of a dominant order in certain languages (Ferrer-i-Cancho 2014).

- *Coexistence*: different word order principles can dominate simultaneously in a language. For instance, SOV languages suggest that the ordering of the triple is determined by the need of minimizing the uncertainty about the verb. Besides, the tendency of these languages to put adjectives before nouns (or auxiliaries after verbs) is a prediction of the principle of dependency length minimization (Ferrer-i-Cancho 2008, Ferrer-i-Cancho 2015a). The converse may happen to SVO languages where the placement of the object is optimal with respect to dependency length minimization but the placement of adjectives before nouns in the nominal constituents could be driven by the principle of minimization of the uncertainty about the head. A beautiful example of coexistence of principles is provided by languages that are not SVO but have SVO as alternative order (Greenberg 1963). There SVO could arise to compensate for a suboptimal choice from the perspective of dependency length minimization (Ferrer-i-Cancho 2015a).
- *Cooperation*: coexistence makes emphasis on the diversity of word orders that may result from conflicting constraints, e.g., a couple of primary alternating orders instead of one (Ferrer-i-Cancho 2016a). The idea of cooperation emphasizes the possibility that two word orders interact to produce the same word order pattern. Take the principle of minimization of the predictability of the head versus dependency length minimization. Suppose that the former beats the latter for the placement of the verb that is then put last. In this context, assuming that the relative placement of adjectives has to be consistent for both the subject and the argument, it follows that dependency length minimization will lead to adjectives before nouns. This is also expected by the principle of head uncertainty minimization for the particular case of nominal heads (Ferrer-i-Cancho 2015a).
- *Neutrality*: certain placements may have literally no clear advantage for the brain with respect to a certain word order principle, e.g., *a priori* adjectives can either follow or precede nouns in SVO languages according to the principle of dependency length minimization (Ferrer-i-Cancho 2015a). Functional pressures do not imply that some orders are better than others in all cases.
- *Word order survival or the recipient of benefits*: certain placements may not be advantageous for the brain with respect to at least one word order principle (they can be neutral as we have seen above). Instead, they could be explained as a result of competition for survival among dominant word orders. For instance, dominant SVO orders may increase their survival by choosing a relative placement of adjectives with respect to nouns that is inconvenient for SOV languages (Ferrer-i-Cancho 2015a). A very important point is that then a certain placement would not be explained by its benefit for the brain but for its benefits for the survival of a word order (Ferrer-i-Cancho 2015b). This represents a radical shift of perspective with respect to the exclusive focus of word order research in cognitive science on benefits for the brain. This hypothesis should be evaluated considering an alternative hypothesis, namely that such a relative placement might be due to the coexistence of a principle to reduce the uncertainty about the nominal heads, that predicts that the nominal head should be put first. However, this alternative is less likely given that heads are normally more costly and thus should be the target.

- *Conditional word order biases or Kauffman's adjacent possible*: word order variation and word order change can be highly determined by the current state of the system (its dominant word order), overriding prior unconditional biases (Ferrer-i-Cancho 2016a).
- Word order evolution as a *path dependent process*: the next steps of word order evolution are determined by history (Dun et al 2011). For instance, running away from the attraction of SOV preadapts SVO languages to become VSO/VOS languages (Ferrer-i-Cancho 2016a). Again, interpreting any word order configuration as arising exclusively from absolute brain costs, as it is customary in cognitive science, can be misguiding. History matters.
- *Symmetry breaking*: to understand word order it is important to understand how the tie between alternative orders could be broken. Some examples are the following:
    - The symmetry between the minimization of the uncertainty about the verb and the uncertainty about its arguments (the nominal constituents) is broken by the fact that verbs are harder to learn.
    - The relative placement of adjectives in SVO languages reviewed above.
    - When the current state is SOV, pressure for dependency length minimizeation predicts two as orders as the most likely: SVO and its symmetric OVS. The word permutation ring hypothesis breaks the symmetry towards SVO.
    - The conflict between dependency length minimization and uncertainty minimization could be broken by the length or the scale in favor of the former. The conflict between principles needs at least three elements and uncertainty minimization starts operating with just two elements (Section 2.2). Then, the former would tend to dominate in longer sequences or at higher scales while the latter would tend to dominate in shorter sequences or at short ranges (Ferrer-i-Cancho 2014, Ferrer-i-Cancho 2015a). Such a division of labour is linked to the concept of coexistence.
- *The shape of the adaptive landscape*: the adaptive landscape of dependency lengths for the case of a single head is quasi-convex under some general assumptions (Ferrer-i-Cancho 2015a). Further research should be carried out to determine if it is also convex and to shed light on the shape of the complex landscape that results when uncertainty minimization is integrated.

The case of Mandarin Chinese can help us to see how the concepts above can be applied. That language has SVO as dominant order and tends to put adjectives before nouns (Dryer & Haspelmath 2013). As we have seen above (*Word order survival or the recipient of benefits*), the dominance and the survival of SVO is enhanced by placing adjectives in a relative position with respect to nouns that is inconvenient for SOV, namely, after nouns. This is not the case of Mandarin Chinese and that may explain the *coexistence* of SOV and SVO in that language (Gao 2008).

The concept of *intrinsic conflicts* and the concept of *collaboration* can be seen as instances of Morin's (1990) dialogic principle, according two principles (dependency length minimization and uncertainty minimization in our case) could be at the same time antagonistic and complementary. This is an example of how the philosophical and episteme-ological approach of "general complexity" can be unified with the mainly scientific and methodological approach of "restricted complexity" (Malaina 2015).

Our theoretical results on the conflict between word order principles provide an answer to the question of the "relative roles of worktye ing memory principles", i.e. dependency length minimization in our framework, "and principles of information theory accounts of sentence processing such as surprisal", i.e. uncertainty minimization/predictability maximization in our setup (Lewis et al 2006). Notice that we do not view information theoretic principles as necessarily external to working memory in our approach.

Early in the article (Section 2), we justified the principles of word order based on extensions of principles that are defined over individual words in optimization principles of communication (Ferrer-i-Cancho 2017a). These lexical principles probably apply beyond lexical elements and then support the "fast content-addressed access to item information" involved in processing of sequences (Lewis et al 2006). It is time to close the circle in the opposite direction. The principle of dependency length minimizeation is a principle of word order that has words as units. Length could be measured with more precision in syllables or phonemes (Ferrer-i-Cancho 2015b). In that way, the length of a dependency would be a function of the length of the words defining the dependency and that of the words in-between. Therefore, word lengths should be minimized to minimize dependency lengths (Ferrer-i-Cancho 2017c). Put differently, dependency length minimization predicts the principle of compression, linking dependency length minimization with the origins of Zipf's law for word frequencies (Ferrer-i-Cancho 2016b, Ferrer-i-Cancho 2017a), Zipf's law of abbreviation (Ferrer-i-Cancho et al 2013b) and Menzerath's law (Gustison et al 2016). Therefore, dependency length minimization predicts reduction, a phenomenon that has been used to justify the uniform information density and related hypotheses (Section 3.4). The need for the need for uniform information density and similar hypotheses as independent standalone hypotheses is seriously challenged. However, we do not mean that dependency length is the only reason for compression. For instance, in small sequences where dependency length minimization is irrelevant or can be neglected (Section 4.5), compression *per se* still matters.

We hope that our sketch of a general theory of word order and beyond stimulates further research. Notice that the scope of the main theoretical results presented above goes beyond linguistics. Uncertainty minimization makes predictions about the optimal placement of target elements for any sequence *a priori*. Dependency length minimiztion requires that there is some structure, e.g., there must be a hub element, the equivalent of a head in a linguistic context. For these reasons our results could be applied to genomic sequences (Searls 1992) or animal behavior sequences (Kershenbaum et al 2016).

# APPENDIX A

*Property*

Suppose that $g_H$ and $g_I$ are two functions whose domain and co-domain are real numbers: $g_H$ is a strictly monotonically increasing function while $g_I$ is a strictly monotonically decreasing function. One has

$$g_H[H(Y|X_1, X_2,\ldots, X_{i-1})] \geq g_H[H(Y|X_1, X_2,\ldots, X_i)] \quad \text{(A.1)}$$

for $i \geq 1$, and

$$g_I[I(Y|X_1, X_2,\ldots, X_{i-1})] \leq g_I[I(Y|X_1, X_2,\ldots, X_i)] \quad \text{(A.2)}$$

for $i \geq 2$.

*Proof*:

One has that

$$H(Y) \geq H(Y|X_1). \tag{A.3}$$

with equality if and only if $Y$ and $X_1$ are independent (Theorem 2.6.5, p. 29, Cover & Thomas 2006). This is obtained by a straightforward application of the fact that "*conditioning reduces entropy*" (in general) or that "*information cannot hurt*" (Cover & Thomas 2006, p. 29).

We would like to prove the general case

$$H(Y|X_1, X_2, ..., X_{i-1}) \geq H(Y|X_1, X_2, ..., X_i) \tag{A.4}$$

for $i > 1$ (the case $i = 1$ corresponds to Eq. A.3). The conditional mutual information between $Y$ and $X_i$ knowing $X_1, \ldots X_{i-1}$ is

$$I(Y; X_i|X_1, X_2, ..., X_{i-1}) = H(Y|X_1, X_2, ..., X_{i-1}) - H(Y|X_1, X_2, ..., X_i) \tag{A.5}$$

Then Eq. A.4 is equivalent to

$$I(Y; X_i|X_1, X_2, ..., X_{i-1}) \geq 0 \tag{A.6}$$

Lemma 3.1 of Wyner (1978) warrants that the inequalities in A.4 and A.6 hold with equality if and only if $X_i$, $X_1, \ldots X_{i-1}$ and $Y$ define a Markov chain.

The properties of $g_H$ give Eq. A.1. A parallel conclusion can be reached for $I(Y; X_1, ..., X_i)$.

Multiplying by -1 in Eq. A.6 one gets

$$-H(Y|X_1, X_2, ..., X_{i-1}) \leq -H(Y|X_1, X_2, ..., X_i), \tag{A.7}$$

Adding $H(Y)$ one gets

$$H(Y) - H(Y|X_1, X_2, ..., X_{i-1}) \leq H(Y) - H(Y|X_1, X_2, ..., X_i) \tag{A.8}$$

and finally Eq. A.2 for $i$ □ 2 (notice that $I(Y; X_1, X_2, ..., X_{i-1})$ is not defined when $i = 1$) as we wanted to prove.

The property above allows one to conclude easily that placing the target last is optimal, namely

$$n \in \text{argmin}_{1\leq i\leq m} g_H[H(Y|X_1, X_2, ..., X_i)] \tag{A.9}$$

and

$$n \in \text{argmax}_{2\leq i\leq m} g_I[I(Y|X_1, X_2, ..., X_i)], \tag{A.10}$$

although not necessarily the only optimum. Therefore, in the absence of any further information, placing the target last is the most conservative strategy and thus it is the optimal in general.

# APPENDIX B

It is easy to show that

$$\text{ma}x_{1\leq i} H(X_i|X_1, X_2, ..., X_{i-1}) = H(X_1), \tag{B.1}$$

assuming Zipf's law. Notice that Hilberg's law (Eq. 11) implies $a = H(X_1)$ and also that Eq. B.1 is approximately equivalent to

$$\max_{1\leq i} a i^{-\gamma} = a. \qquad \text{(B.2)}$$

as $\gamma$ is strictly positive.

## Acknowledgements

The present article is an evolved version of some of the arguments of the unpublished manuscript "Optimal placement of heads: a conflict between predictability and memory". The major result of the article was the conflict between the principle dependency length minimization and predictability maximization. The article was submitted for publication in September 2009 after being presented as "Memory versus predictability in syntactic dependencies" in the Kickoff Meeting "Linguistic Networks" (Bielefeld University, Germany) in June 5, 2009. We thank the participants of the Kickoff Meeting, specially G. Heyer and A. Mehler for valuable discussions. A more advanced version was presented in 2011 as "Word order as a constraint satisfaction problem: A mathematical approach." In the workshop "Complexity in Language: Developmental and Evolutionary Perspectives" (Collegium de Lyon, May 23 - 24).

Since 2009, at least R. Levy, F. Jaeger, E. Gibson, S. Piantadosi and R. Futrell have had access to different versions of the unpublished manuscript. Evolved versions of various components of the unpublished manuscript have already appeared (Ferrer-i-Cancho 2014, 2015a).

For the present article, we thank C. Bent and S. Semple for their careful revision. We are also grateful to Ł. Dębowski, G. Fenk-Oczlon, F. Moscoso del Prado Martín, M. Wang, and Eric Wheeler for helpful comments and discussions, to S. Wichmann for pointing us to Hammarström's work, and to Y. N. Kenett for pointing us to Cowan's work. This research was funded by the grants 2014SGR 890 (MACDA) from AGAUR (Generalitat de Catalunya) and also the APCOM project (TIN2014-57226-P) from MINECO (Ministerio de Economia y Competitividad).